\documentclass[conference]{IEEEtran}
\IEEEoverridecommandlockouts
\usepackage{cite}
\usepackage{amsmath,amssymb,amsfonts}
\usepackage{algorithmic}
\usepackage{graphicx}
\usepackage{textcomp}
\usepackage{multirow}
\usepackage{float}
\usepackage{caption}
\usepackage{subcaption}
\usepackage{mathtools}
\usepackage{colortbl}
\usepackage{balance}

\newcommand{\ANNColor}{ffe2bd}
\newcommand{\SNNColor}{a8d8ff}
\usepackage[table,xcdraw]{xcolor}

\def\BibTeX{{\rm B\kern-.05em{\sc i\kern-.025em b}\kern-.08em
    T\kern-.1667em\lower.7ex\hbox{E}\kern-.125emX}}
\begin{document}

\title{Towards Efficient Deployment of Hybrid SNNs on Neuromorphic and Edge AI Hardware}

\author{\IEEEauthorblockN{James Seekings, Peyton Chandarana, Mahsa Ardakani, MohammadReza Mohammadi, and Ramtin Zand}
\IEEEauthorblockA{Department of Computer Science and Engineering, University of South Carolina, Columbia, USA \\
seekingj@email.sc.edu, psc@email.sc.edu, mahsam@email.sc.edu, mohammm@email.sc.edu, ramtin@cse.sc.edu }
}

\maketitle

\begin{abstract} 
This paper explores the synergistic potential of neuromorphic and edge computing to create a versatile machine learning (ML) system tailored for processing data captured by dynamic vision sensors. We construct and train hybrid models, blending spiking neural networks (SNNs) and artificial neural networks (ANNs) using PyTorch and Lava frameworks. Our hybrid architecture integrates an SNN for temporal feature extraction and an ANN for classification. We delve into the challenges of deploying such hybrid structures on hardware. Specifically, we deploy individual components on Intel's Neuromorphic Processor Loihi (for SNN) and Jetson Nano (for ANN). We also propose an accumulator circuit to transfer data from the spiking to the non-spiking domain. Furthermore, we conduct comprehensive performance analyses of hybrid SNN-ANN models on a heterogeneous system of neuromorphic and edge AI hardware, evaluating accuracy, latency, power, and energy consumption. Our findings demonstrate that the hybrid spiking networks surpass the baseline ANN model across all metrics and outperform the baseline SNN model in accuracy and latency.

%Spiking Neural Networks (SNNs) are a form of event-based neural network that offers low-power computation by employing biologically inspired brain neural dynamics. These networks are especially effective at extracting temporal features from event-based data due to their execution over time. However, SNNs often underperform in classification compared to other Artificial Neural Networks (ANNs) which use dense and highly parallel operations. These operations are costly due to their floating point or integer computations vastly increasing both power consumption and execution time which is not ideal when considering large-scale networks. Thus,  leveraging the strengths of both types of networks in a hybrid SNN and ANN system, which we call a Hybrid Spiking Neural Network (HSNN), could yield in a system which is both accurate and energy efficient.
%which we call a Hybrid Spiking Network. 
%This proposed network will use an SNN backbone for temporal feature extraction and an ANN head for classification. The individual components are deployed on Intel's Neuromorphic Processor Loihi (SNN) and Jetson Nano (ANN) to profile their performance. Our results show that the hybrid spiking networks outperform the baseline ANN model in every metric and the baseline SNN model in terms of accuracy and latency.

%achieve better accuracy compared to the baseline SNN model with comparable energy consumption    

%achieves accuracy on par with baseline ANNs while maintaining energy efficiency similar to baseline SNNs.  

\end{abstract}


\section{Introduction}
\label{sec:intro}

Spiking Neural Networks (SNNs) \cite{lobo2020spiking,ghosh2009spiking} are an emerging technology aimed at creating biologically-inspired neural networks for low-power and high-performance computation. They utilize neurons modeled after the brain, enabling them to learn over time and excel at extracting temporal information from event-based data \cite{Loihisurvery,eshraghian2023training,LessonsLoihi}. In contrast, artificial neural networks (ANNs) like convolutional neural networks (CNNs) are proficient at extracting spatial information but do not handle temporal information well \cite{li2021survey}. Recurrent neural networks (RNNs) have gained popularity for their ability to handle temporal information, but they do not offer significant improvements in terms of power or latency \cite{sherstinsky2020fundamentals}. 

SNNs are being explored as a promising alternative for conventional ANNs due to their low-power computing capabilities, yet when deployed on existing neuromorphic hardware, they often underperform in terms of classification accuracy \cite{Mohammadi_2022, chandarana_igsc_2022, smithICMLA2023}. One solution to benefit from the advantage of both SNN and ANN models is to fuse them to create more robust and versatile neural network models capable of addressing complex tasks, including pattern recognition in time-series data and understanding dynamic systems in real-time applications. However, integrating SNNs and ANNs in a single system presents challenges such as developing efficient training algorithms to train across the two domains and optimizing the use of computational resources. Despite these challenges, exploring the viability of combining ANNs and SNNs into one system could yield promising results to advance neural networks capabilities. %and foster innovation in neural network research.

In 2021, Kugele et al. \cite{kugele_hybrid_snn_2021} proposed a hybrid SNN-ANN architecture with a custom simulator to compile and train a hybrid neural network. Their proposed model consists of an SNN backbone for extracting temporal information and an ANN head for classification. 
%Additionally, we investigate the ha were then able to take these models and measure their power, energy, and latency to test all around efficiency. 
Inspired by \cite{kugele_hybrid_snn_2021}, Wu et al. \cite{wu_hardware_aware_2023} investigates appending dense layers to SNN networks to improve accuracy on CIFAR-10 \cite{Krizhevsky09learningmultiple}. In \cite{kosta_live_demo_2023}, a hybrid SNN-ANN model is utilized to process the data captured by dynamic vision sensor (DVS) \cite{DVS1, DVS2}. Instead of using the SNN for the backbone, other works such as Muramatsu et. al. \cite{muramatsu_combining_snn_2021} explore using the ANN as a backbone with an SNN head for classification on the MNIST \cite{lecun2010mnist} and CIFAR-10 datasets. Additionally, Muramatsu et. al. performed multiple experiments by modifying the ratios of ANN to SNN layers concluding that models with more ANN layers typically achieve better accuracy. Beyond the domain of image classification, a few papers \cite{cordone_object_detection_2022,zhang2023direct} have also explored the use of hybrid networks in object detection tasks.  This is done by combining an SNN backbone with a single-shot detector head and using a surrogate gradient to train the networks. %The automotive datasets they used are event-based. 

Herein, our work provides a deeper investigation of hybrid SNN-ANN models by offering a unified training mechanism that operates across the SNN and ANN domains. Additionally, we undertake a more extensive performance analysis for different architectures of the hybrid network. %Furthermore, we extend the research proposed in Kugele et. al. \cite{kugele_hybrid_snn_2021} improving upon the training of hybrid models using backpropagation via a unified SLAYER-PyTorch training pipeline which produces trained SNN and ANN models that are readily deployable on their respective hardware platforms.
The main contributions of our paper compared to the previous works are:
\begin{itemize}
    \item Developing a unified backpropagation-based training mechanism for hybrid spiking and non-spiking architectures using PyTorch and SLAYER \cite{shrestha2018slayer} as part of the LAVA neuromorphic computing library \cite{LAVA}.% SLAYER \cite{shrestha2018slayer} is a framework that allows for evaluating the gradients of various SNN models via a novel temporal credit assignment policy. 
    \item Investigating the hardware deployment challenges of hybrid architectures. This paper is one of the pioneers in attempting the real hardware implementation of hybrid SNN-ANN models.
    \item Providing comprehensive performance analyses of hybrid SNN-ANN models deployed on a heterogenous system of neuromorphic and edge AI hardware. %in terms of accuracy, latency, power, and energy consumption on a hetneuromorphic hardware.
\end{itemize}

The remainder of the paper is organized as follows. Section \ref{sec:hybrid-arch} introduces the proposed hybrid SNN-ANN architectures and their training mechanism. Section \ref{sec:deployment} demonstrates the hardware deployment of our hybrid architecture on a heterogeneous system of neuromorphic hardware and edge AI accelerators. The experiments performed and the results obtained are discussed in Section \ref{sec:results}  . Finally, Section \ref{sec:conclusion} concludes the paper.

\section{Proposed Hybrid SNN-ANN Architecture}
\label{sec:hybrid-arch}

\begin{figure}[]
\begin{subfigure}[b]{3.45in}
    \centering
    \includegraphics[width=\linewidth]{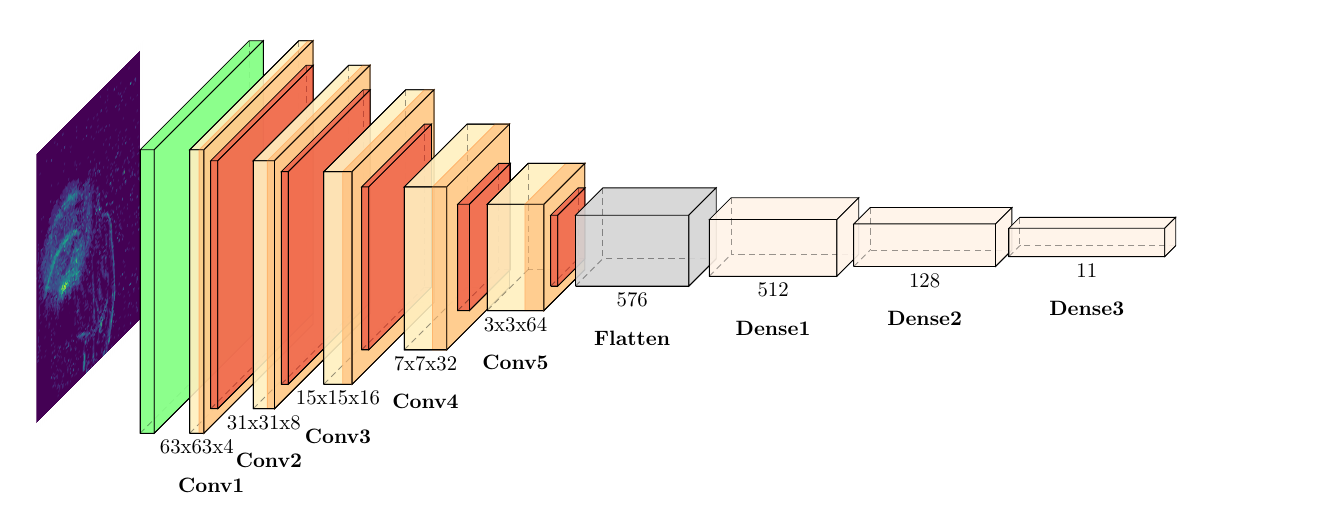}
    %\vspace{-2mm}
    \caption{ANN Architecture}
    %\vspace{-11mm}
    \label{fig:ann_arch}
\end{subfigure}
\begin{subfigure}[b]{3.45in}
    
    \centering
    \includegraphics[width=\linewidth]{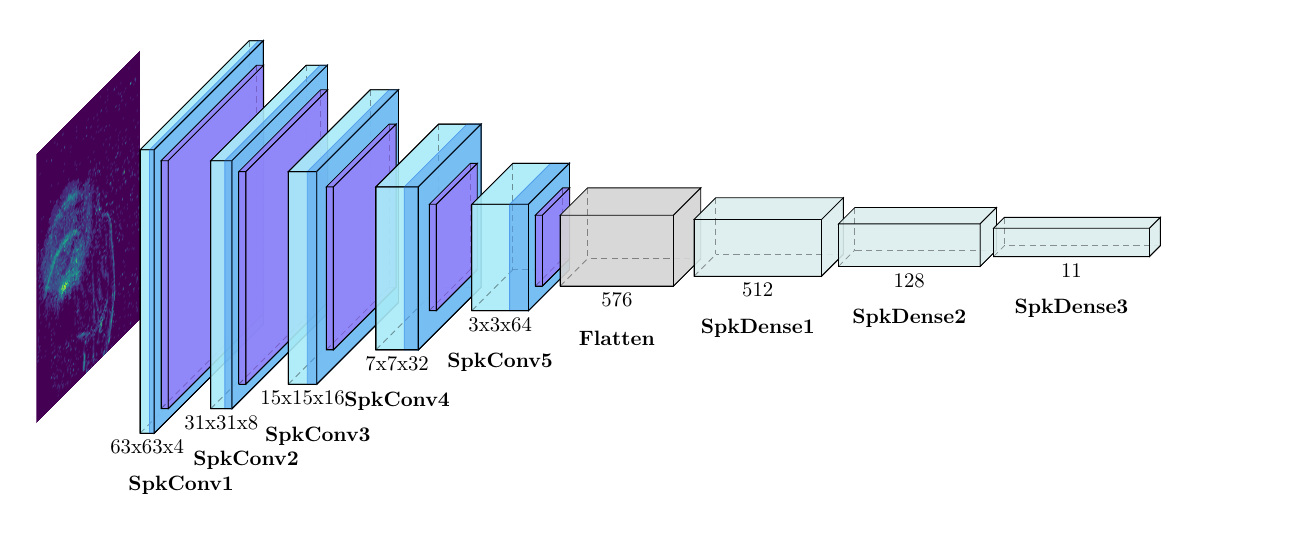}
    %\vspace{-2mm}
    \caption{SNN Architecture}
    %\vspace{-11mm}
    \label{fig:snn_arch}
\end{subfigure}
\begin{subfigure}[b]{3.45in}

    \centering
    \includegraphics[width=\linewidth]{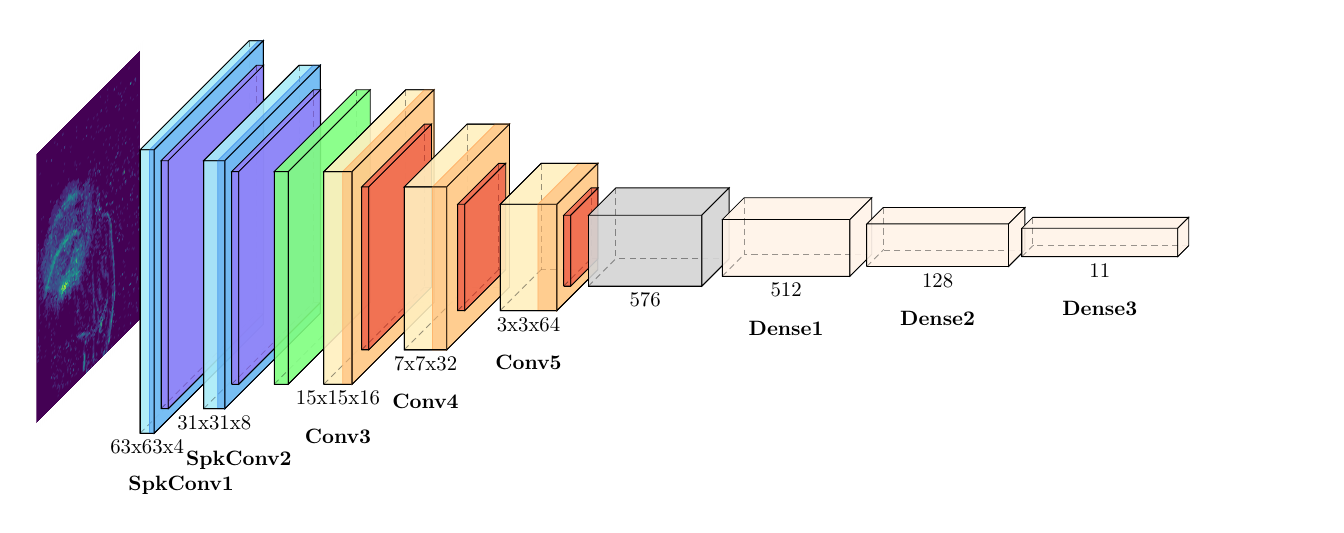}
    %\vspace{-2mm}
    \caption{Example $S_2A_3$ Hybrid Architecture}
    %\vspace{-11mm}
    \label{fig:hnn_arch}
\end{subfigure}

\caption{The architecture of the CNN models investigated herein to process the data captured by the DVS camera.}
\label{fig:architecture}
\end{figure}

\subsection{Hybrid Architecture}

Here, we develop a representative CNN architecture with five convolution layers and three dense layers to test the effectiveness of integrating spiking and non-spiking components. Figure \ref{fig:ann_arch} shows the baseline ANN architecture using non-spiking convolution and dense blocks that employ ReLU activations and max pooling operations. Additionally, an accumulate operation is included at the front of the network, represented by the green layer, which collapses the temporal dimension from the input data so that the ANN can process it. Figure \ref{fig:snn_arch} shows the baseline SNN architecture consisting of Spike Convolutions (SpkConv) and Spike Dense (SpkDense) blocks. These blocks use Current-Based Leaky Integrate and Fire (CUBA-LIF) neurons as activations and spike pooling operations. 

A hybrid network is generated by replacing the Conv layers from the ANN with SpkConv layers from the SNN. Layers are replaced in order starting from the beginning of the model and moving deeper. The accumulate operation is then placed between the spiking and non-spiking layers to remove the temporal dimension as the data is passed to ANN, and the dense layers are always implemented via non-spiking blocks. Figure \ref{fig:hnn_arch} shows an example of a hybrid network with two SpkConv layers and three regular Conv layers, labeled $S_2A_3$. Table \ref{tab:hybrid-archs} provides further details of the model architectures investigated in this paper.

\begin{table*}[h!]
\centering
\caption{Network Architectures. The spiking and non-spiking layers are indicated by ``Spk'' and ``non-Spk'', respectively. Accumulate interval is denoted with $I=(5/10/25)$.}
\label{tab:hybrid-archs}
\begin{tabular}{cccccccccccc}
\hline
& \multicolumn{2}{c}{\textbf{\begin{tabular}[c]{@{}c@{}}Convolutions\end{tabular}}} && \multicolumn{2}{c}{\textbf{\begin{tabular}[c]{@{}c@{}}Linear\end{tabular}}} && \multicolumn{3}{c}{\textbf{Parameters}} & &  \\ 
\cline{2-3} \cline{5-6} \cline{8-10}
\textbf{\begin{tabular}[c]{@{}c@{}}Model\end{tabular}} & 
\textbf{\begin{tabular}[c]{@{}c@{}}Spk\end{tabular}} & \textbf{\begin{tabular}[c]{@{}c@{}}Non-Spk\end{tabular}} && 
\textbf{\begin{tabular}[c]{@{}c@{}}Spk\end{tabular}} & \textbf{\begin{tabular}[c]{@{}c@{}}Non-Spk\end{tabular}} &&
$I=5$ & $I=10$ & $I=25$ && \textbf{Loihi Cores} \\ \hline

\cellcolor[HTML]{\ANNColor}ANN  & 
\cellcolor[HTML]{\ANNColor}0 & \cellcolor[HTML]{\ANNColor}5  & \cellcolor[HTML]{\ANNColor}{} & 
\cellcolor[HTML]{\ANNColor}0 & \cellcolor[HTML]{\ANNColor}3  & \cellcolor[HTML]{\ANNColor}{} & 
\cellcolor[HTML]{\ANNColor}223,991 & \cellcolor[HTML]{\ANNColor}223,631  & \cellcolor[HTML]{\ANNColor}223,415 &
\cellcolor[HTML]{\ANNColor} & \cellcolor[HTML]{\ANNColor}0 \\ \hline
 
$S_1A_4$ & 1 & 4 && 0 & 3 && 225,943 & 224,503 & 223,639 && 16 \\ \hline
$S_2A_3$ & 2 & 3 && 0 & 3 && 233,727 & 227,967 & 224,511 && 32 \\ \hline
$S_3A_2$ & 3 & 2 && 0 & 3 && 264,839 & 241,799 & 227,975 && 36 \\ \hline
$S_4A_1$ & 4 & 1 && 0 & 3 && 389,263 & 297,103 & 241,807 && 38 \\ \hline
$S_5A_0$ & 5 & 0 && 0 & 3 && 3,041,431 & 1,566,871 & 683,135 && 42 \\ \hline

\cellcolor[HTML]{\SNNColor}SNN  & 
\cellcolor[HTML]{\SNNColor}5 & \cellcolor[HTML]{\SNNColor}0 & \cellcolor[HTML]{\SNNColor} & 
\cellcolor[HTML]{\SNNColor}3 & \cellcolor[HTML]{\SNNColor}0 & \cellcolor[HTML]{\SNNColor} & 
\cellcolor[HTML]{\SNNColor}387,229 & \cellcolor[HTML]{\SNNColor}387,229  & \cellcolor[HTML]{\SNNColor}387,229 &
\cellcolor[HTML]{\SNNColor} & \cellcolor[HTML]{\SNNColor}58 \\ \hline

\end{tabular}%
\end{table*}

\subsection{Accumulator}

Before event-based data can be passed to ANN, the temporal dimension must be collapsed. To accomplish this, methods such as those used in \cite{kugele_hybrid_snn_2021} involve implementing an accumulator that sums spikes over small time intervals to generate multiple outputs to the ANN. Each output is run through the first layer of the ANN and then concatenated together at the second layer. Our accumulator differs in that, after summation, the outputs are concatenated together before being sent to ANN. The concatenation occurs across the channel dimension, causing it to expand with the size of the temporal dimension. 

Herein, the period over which the accumulator sums spikes is referred to as the \textit{accumulate interval}. A large interval sums up many spikes at once, reducing output size at the cost of temporal resolution. On the other hand, smaller intervals better retain temporal resolution but lead to increased model size. In our experiments, we treat the accumulate interval as a hyperparameter to investigate its effect on model performance, which we discuss in Section \ref{sec:results}. A model's accumulate interval is denoted by $I=(5/10/25)$, as shown in Table \ref{tab:hybrid-archs}.

\begin{figure}[h!]
    \centering
    \includegraphics[width=3.45in]{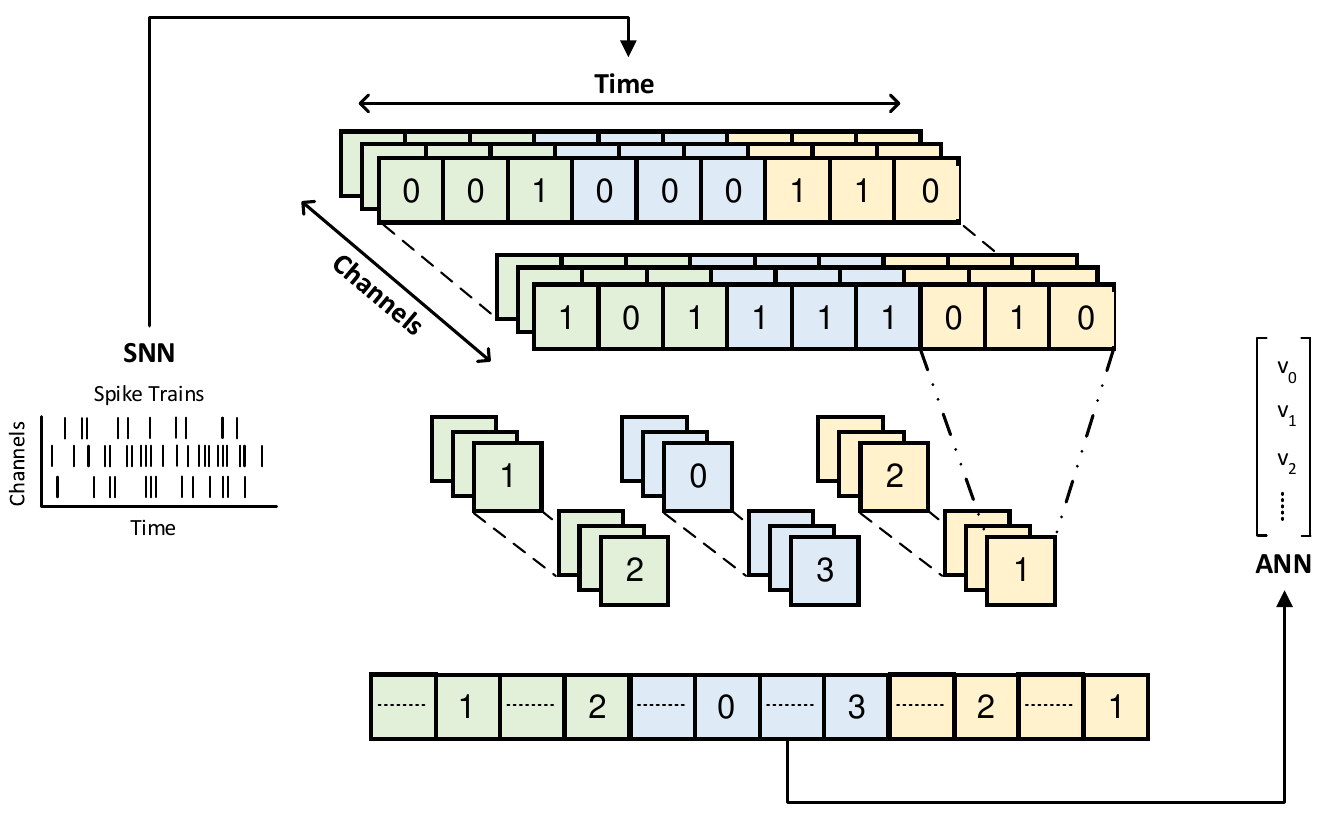}
    %\vspace{-2mm}
    \caption{An example of accumulate operation on a 2-dimensional data with 9 timesteps (T=9) and an accumulate interval of 3 (I=3).}
    %\vspace{-11mm}
    \label{fig:accumulator}
\end{figure}

Figure \ref{fig:accumulator} shows the accumulate operation performed on 2-dimensional data although it can be generalized for 4-dimensional input. First, the input data is separated into groups of size $I$ along the temporal dimension, represented by different colors. The groups are then summed together resulting in channel-wide vectors. After summation, these vectors are concatenated together with the earlier elements appearing first, preserving temporal order. The output of the accumulator is then sent to ANN for further processing.

A mathematical representation of the accumulate operation can be seen in Eq. \ref{eq:forwards_pass}, where $S$ is a $C \times T$ matrix representing spiking input. $C$ is the number of channels and $T$ is the number of time steps. The output is a vector $A$ with length $CT/I$, where $I$ is the accumulate interval.

\begin{equation}
A_{j} = \sum_{k=0}^{I-1} S_{(j\ mod\ C),(I\lfloor\frac{j}{C}\rfloor+k)}
\label{eq:forwards_pass}
\end{equation}

\noindent Each output $A_j$ is defined as the summation of $S$ at channel $j\ mod\ C$ over $I$ timesteps. Once $j$ exceeds $C$, the summation resets to the first channel through the modulus operation. At this point, $\lfloor\frac{j}{C}\rfloor$ equals 1 which shifts the columns for summation over by $I$. This repeats for every $C$ indices until the final index. 

One challenge of this method is that the channel dimension expands at a rate of $T/I$ to accommodate the shrinking temporal dimension. Reducing the channel count back down to what it was before the accumulate operation can be done with a simple convolutional layer. However, this is not possible in models such as $S_5A_0$ which does not have a convolution layer following the accumulate operation. As such, those models experience a substantial increase in parameter count at smaller accumulate intervals, as seen in Table \ref{tab:hybrid-archs}.

\subsection{Training} 

Our hybrid model was built in PyTorch using the LAVA framework \cite{LAVA} for spiking components. The LAVA library contains a SLAYER-based training algorithm for SNNs which saves the network's previous states to be used during the calculation of gradients via a temporal credit assignment policy \cite{shrestha2018slayer} and Back Propagation Through Time (BPTT) \cite{NEURIPS2018_82f2b308}. %This process allows the gradients to be shared with respect to time in a method called Back Propagation Through Time (BPTT) \cite{NEURIPS2018_82f2b308}. 
The training algorithms for ANN and SNN build computational graphs for calculating gradients and the graphs are combined automatically, allowing the hybrid model to train as a single unified network instead of being trained separately. However, the accumulate operation is non-differentiable, requiring a custom backward pass to be implemented.

In the backward pass of the accumulator, we are faced with an inverse of the forward pass challenge encountered previously. Here, 3-dimensional gradients are received from the ANN while the SNN expects 4-dimensional gradients, thus the temporal dimension needs to be reintroduced or expanded from the ANN domain. This is done by repeating the gradients in-place $I$ times and then reshaping the data to four dimensions. Through this process, spikes that were initially summed together share the same gradient. A mathematical representation following the logic of the forward pass can be seen in Eq. \ref{eq:backwards_pass}.

\begin{equation}
S_{i,j} = A_{C\lfloor \frac{j}{I} \rfloor+i}
\label{eq:backwards_pass}
\end{equation}

\section{Deployment Methodology} 
\label{sec:deployment}

\begin{figure}[]
    \centering
    \includegraphics[width=3.45in]{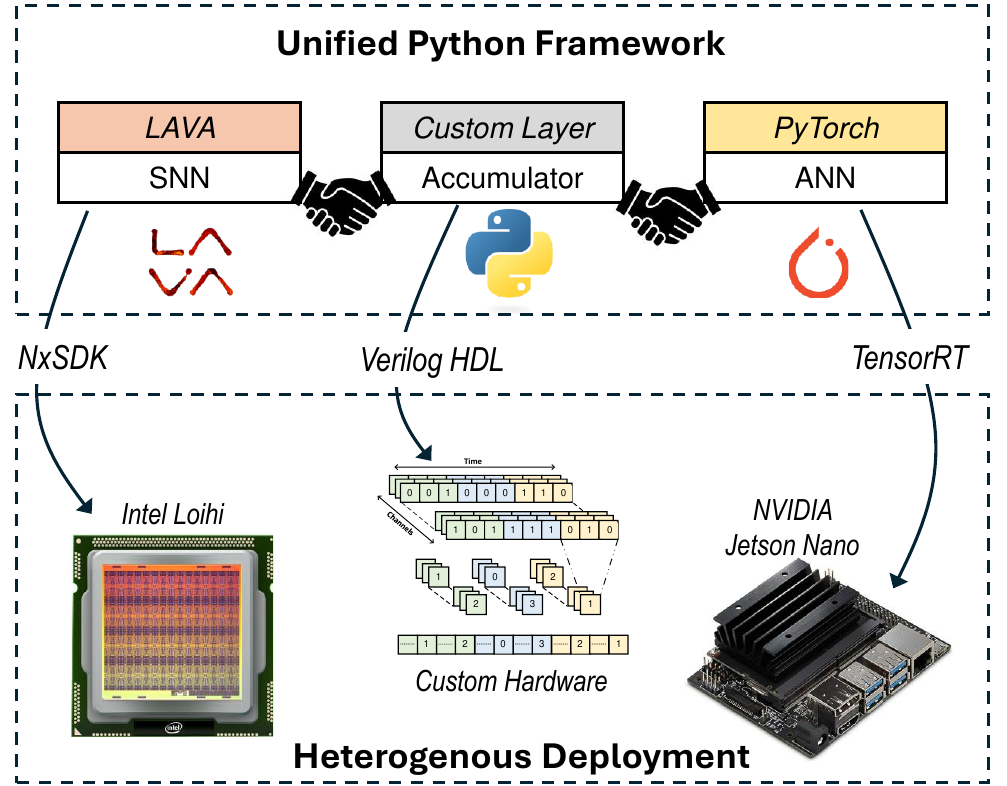}
    %\vspace{-2mm}
    \caption{The proposed unified python framework for training the hybrid SNN-ANN models and the corresponding deployment on a heterogenous system of neuromorphic hardware and edge AI accelerators.}
    %\vspace{-11mm}
    \label{fig:deploy}
\end{figure}

Figure \ref{fig:deploy} shows the end-to-end system in software and hardware starting from the training phase using our unified training pipeline to deploying the networks on their respective hardware. As shown, the spiking and non-spiking components of our proposed hybrid network have differing hardware constraints that limit deployment options. Due to their asynchronous event-based nature, SNNs cannot be run on GPUs or CPUs without simulation which increases latency and consumes more power. Neuromorphic hardware such as Intel's Loihi chip \cite{loihi_2018} is specially designed for running spiking models by emulating biologically inspired neuron dynamics in hardware. However, these chips do not support all of the operations in the ANN models making them unfit to run ANNs. In recent years, there has been various research on the deployment of ANN models on edge AI accelerators \cite{reidy2023efficient,edge_FER,Caveline}. As shown in Fig. \ref{fig:deploy}, we chose to deploy our hybrid spiking model through a distributed system combining a Loihi chip and a Jetson Nano \cite{Jetson-Nano}. These devices were profiled separately to isolate their specific impact on the overall system.

\subsubsection{Deployment of Non-Spiking Components on Edge AI Accelerator}

The NVIDIA Jetson Nano is a development board tailored for ML applications. It utilizes the Tegra X1 System on Chip (SoC), which includes a quad-core ARM Cortex A57 CPU clocked at 1.43 GHz. Additionally, the board features four discrete processing clusters, each with 32 CUDA cores, totaling 128 CUDA cores based on the Maxwell architecture. Equipped with 4 GB of RAM, the Jetson Nano provides a robust computational platform for ML acceleration at the edge.

The Jetson Nano operates in two distinct power configurations: a low-power 5 W mode and a higher-performance Max-N mode, both selectable via a software interface. In the 5 W mode, the device restricts itself to utilizing only two ARM A57 cores at a reduced frequency of 0.9 GHz, and the GPU operates at a reduced clock speed of 0.64 GHz. Conversely, the Max-N mode enables all four ARM A57 cores, running them at an increased frequency of 1.5 GHz, while the GPU operates at a speed of 0.92 GHz. This dual-mode functionality allows developers to balance between power efficiency and computational performance based on the needs of their applications.

Although Jetson Nano is equipped with CUDA capabilities, it primarily utilizes NVIDIA TensorRT \cite{TensorRT} to optimize and accelerate deep learning models through quantization and other optimization techniques, thereby enhancing performance. Models can be exported to TensorRT through the Open Neural Network Exchange (ONNX) \cite{ONNX} library. 

The proposed hybrid networks were recreated in TensorFlow \cite{tensorflow2015-whitepaper} with the spiking components removed and the input layers adjusted. These partial models were then converted to ONNX and exported to TensorRT for deployment on the Jetson Nano. For inference latency, we calculate the time required for 100 inferences and then determine the average inference time for a single input sample. Input, CPU, and GPU power dissipation were recorded through the Jetson Nano's built-in sensors while running each model for three minutes. The \textit{tegrastats} utility is used to read the sensors automatically \cite{Jetson-Nano}. 

\subsubsection{Deployment of Spiking Components on Neuromorphic Hardware}

SNN simulators such as INIsim \cite{rueckauer2017conversion} or Brian 2 \cite{stimberg2019brian} can be used to run SNNs on the CPU through a software abstraction of neural dynamics consuming considerable power and increasing execution latency. Devices that instead emulate neural dynamics, allowing for efficient execution of spiking networks, are referred to as neuromorphic hardware platforms. In our paper, we utilize Intel's Loihi \cite{loihi_2018}, to deploy the spiking components of our hybrid network. 

The Loihi chip departs from the typical von Neumann architecture to more closely replicate the neural dynamics of the brain. It contains a network of neurons and synapses that communicate asynchronously through discrete spike events for more efficient computation. Loihi is organized into 128 programmable cores, each containing 1024 neurons connected by a total of 128 million synapses. While not yet commercially available, 
%our lab has access to the Loihi chip through our partnership with 
access to the Loihi chip is provided to us by Intel Labs through membership in the Intel Neuromorphic Research Community (INRC).

%...\textcolor{orange}{(Add information about how specifically we got access to the chip)}. 

While the LAVA framework is utilized in this paper to train the hybrid and spiking networks, we could not measure the power reliably using the power measurement tools available in LAVA at the time of conducting this research. Consequently, we opted for deploying the models on Loihi 1 and employed the NengoLoihi toolchain \cite{bekolay2014nengo} for deployment and power measurements. NengoLoihi facilitates the conversion of CNNs into SNN architectures through a mapping algorithm, which maps the weights and activations of the CNN onto an equivalent SNN in Intel's NxSDK framework.

%While LAVA framework is used in this paper to train the hybrid and spiking networks, at the time of conducting the research the power measurement tools available in LAVA could not provide reliable results for the model sized used in our experiment. Therefore, we opted for deploying the models on Loihi 2 and using NengoLoihi toolchain \cite{bekolay2014nengo} for deployment and power measurements. Through the NengoLoihi toolchain, it is possible to convert CNNs into SNN architectures through a mapping algorithm that maps the weights and activations of the CNN onto an equivalent SNN from the Intel's NxSDK framework.  

%Deployment on Loihi requires a specific network architecture to be defined using Intel's NxSDK framework, which differs from the LAVA framework we used to train our hybrid models. However, through the NengoLoihi toolchain \cite{bekolay2014nengo}, it is possible to convert CNNs into SNNs with the correct architecture. This is done through a mapping algorithm that maps the weights and activations of the CNN onto an equivalent SNN from the NxSDK framework. 

These initial CNNs were first developed in PyTorch and then mapped to the spiking domain of each hybrid network. Once converted into SNNs through NengoLoihi, each network is partitioned layer-by-layer across Loihi's neuro-cores. After which, neurons and synapses are mapped together between the chip and network. Once partitioned and mapped, the SNN is run directly on the chip. The power dissipation was recorded through a profiling toolkit in NengoLoihi which also reports the number of Loihi cores allocated to the network. The last column of Table \ref{tab:hybrid-archs} provides the number of Loihi cores for all the hybrid and SNN architectures studied herein. As listed, one Loihi chip is sufficient to support the implementation of our hybrid SNN-ANN and SNN models. The latency is measured via the spike propagation delay defined as the time between an image being exposed to the network and the output of spikes from the final layer.

\section{Experiments and Results}
\label{sec:results}

\subsection{Dataset}

Throughout our experiments, we use the DvsGesture dataset which includes 11 hand gestures recorded from 29 subjects under 3 illumination conditions \cite{Amir_2017_CVPR}. %The data is recorded using a DVS which detects the changes in brightness of pixels. 
DVS events are defined by their type (on/off), pixel location (x,y), and a timestamp. To transform the raw DVS data into usable training data, all of the events that occurred in a 10ms time frame were compiled into a 128$\times$128 image. We then took 50 consecutive images to form a single sample of shape (2, 128, 128, 50), representing 500ms of activity. The raw data includes the times when certain gestures are made, which is then used to automatically label the samples. This generates 14,672 training samples and 3,793 testing samples.

\subsection{Accuracy}
\begin{figure}[]
\centering
\includegraphics[width=3.45in]{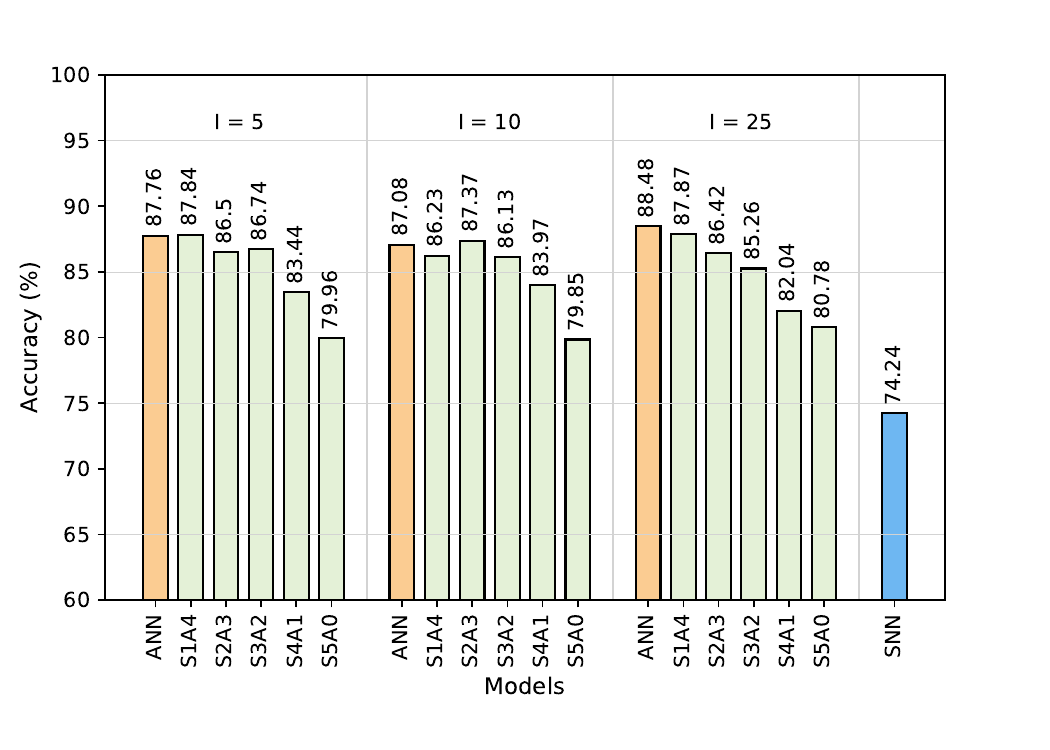}
\caption{Comparative analysis of accuracy for various ANN, SNN, and hybrid SNN-ANN models.} 
\label{fig:accuracy}
\end{figure}

Figure \ref{fig:accuracy} provides a comparison of the accuracies between baseline ANN and SNN, as well as the hybrid networks. To determine the impact on model performance, the models are evaluated with three distinct accumulation intervals: I=5, I=10, and I=25. The data prominently shows that ANNs maintain high accuracy levels across all intervals, achieving their highest at 88.48\% with I=25. On the other hand, the SNN baseline significantly underperforms at 74.24\%. Analyzing the hybrid network results, we see that the continued addition of SpkConv layers leads to accuracy loss at all intervals, mirroring the SNN performance. However, the inclusion of a few SpkConv layers does not have a significant impact on accuracy and in select cases, such as the $S_2A_3$ (I=10) configuration, can even lead to slight increases in accuracy.

\begin{figure}
    \centering
    \includegraphics[width=1.0\linewidth]{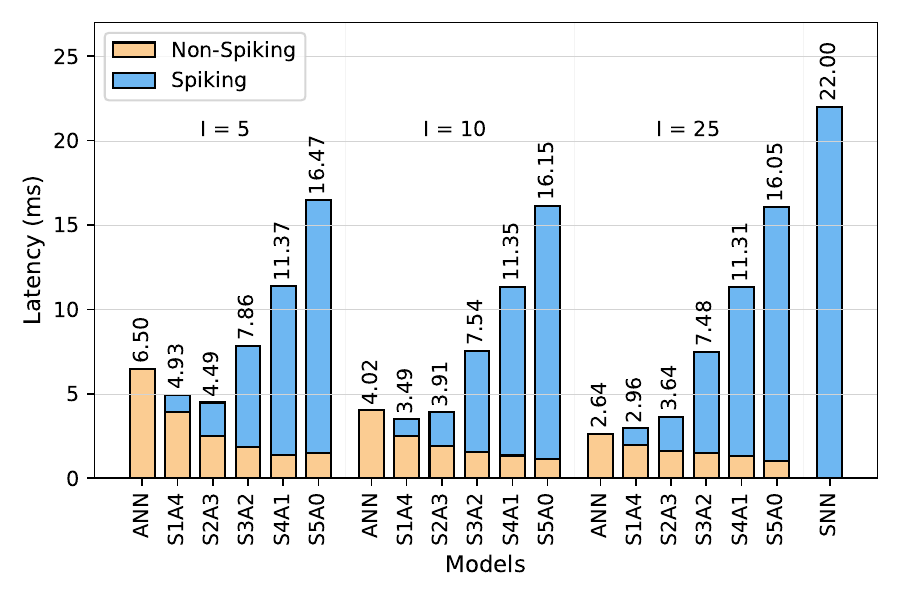}
    \caption{Inference latency for various ANN, SNN, and hybrid SNN-ANN models.}
    \label{fig:latency}
\end{figure}

\subsection{Latency}
Figure \ref{fig:latency} shows the latency of each model, separated for spiking and non-spiking components. The ANN baseline shows 3.28$\times$, 5.47$\times$, and 8.33$\times$ less latency compared to the SNN baseline for I=5, I=10, and I=15, respectively. As shown in the figure, the accumulate interval has a considerable effect on latency, ranging from 6.5 ms to 2.64 ms as it increases. Similar to the accuracy results, it can be observed that the continued addition of SpkConv layers in hybrid networks increases latency to eventually match the SNN's performance. However, a small decrease in latency can be seen when only a few SpkConv layers are present with intervals I=5, and I=10. 

\begin{figure}[]
\centering
\includegraphics[width=1.0\linewidth]{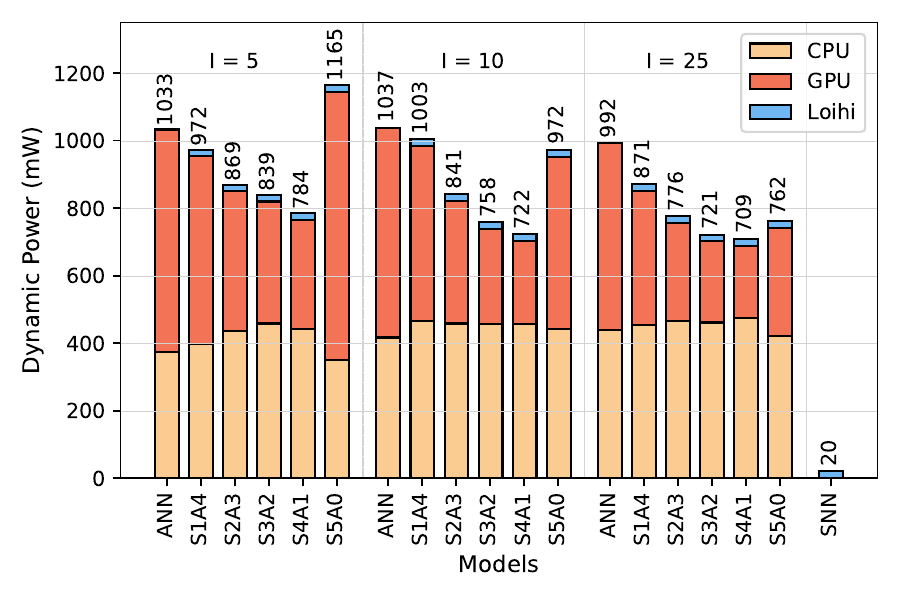}
\caption{Dynamic Power Comparison for CPU, GPU, and Loihi for various ANN, SNN, and hybrid SNN-ANN models.}
\label{fig:dynamic-power}
\end{figure}

\begin{figure}[]
    \centering
\includegraphics[width=1.0\linewidth]{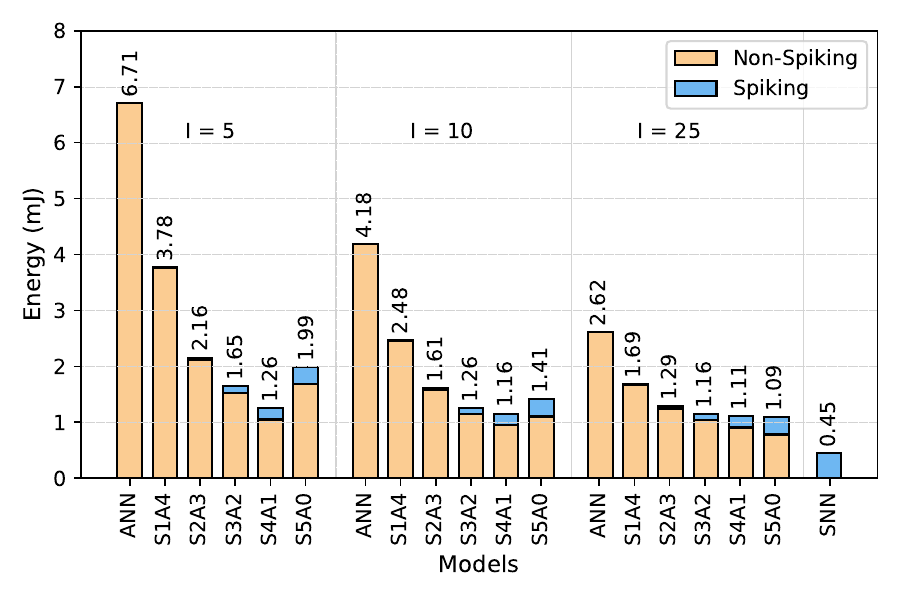}
\caption{Energy Consumption Comparison for various ANN, SNN, and hybrid SNN-ANN models.}
\label{fig:energy}
\end{figure}

\begin{table*}[]
\centering
\caption{Accumulator Performance Results. Recorded using Design Compiler.}
\label{tab:accumulator-results}
\begin{tabular}{cccccccccccc}
\hline
& \multicolumn{3}{c}{\textbf{\begin{tabular}[c]{@{}c@{}}$I=5$\end{tabular}}} && \multicolumn{3}{c}{\textbf{\begin{tabular}[c]{@{}c@{}}$I=10$\end{tabular}}} && \multicolumn{3}{c}{\textbf{$I=25$}}  \\ 
\cline{2-4} \cline{6-8} \cline{10-12}
\textbf{\begin{tabular}[c]{@{}c@{}}Model\end{tabular}} & 
\textbf{\begin{tabular}[c]{@{}c@{}}Power\\(mW)\end{tabular}} & \textbf{\begin{tabular}[c]{@{}c@{}}Latency\\(ms)\end{tabular}} & \textbf{\begin{tabular}[c]{@{}c@{}}Energy \\(mJ)\end{tabular}} && 
\textbf{\begin{tabular}[c]{@{}c@{}}Power\\(mW)\end{tabular}} & \textbf{\begin{tabular}[c]{@{}c@{}}Latency\\(ms)\end{tabular}} & \textbf{\begin{tabular}[c]{@{}c@{}}Energy\\(mJ)\end{tabular}} &&
\textbf{\begin{tabular}[c]{@{}c@{}}Power\\(mW)\end{tabular}} & \textbf{\begin{tabular}[c]{@{}c@{}}Latency\\(ms)\end{tabular}} & \textbf{\begin{tabular}[c]{@{}c@{}}Energy\\(mJ)\end{tabular}} \\ \hline

\cellcolor[HTML]{\ANNColor}ANN & 
\cellcolor[HTML]{\ANNColor}0.294 & \cellcolor[HTML]{\ANNColor}1.28e-2 & \cellcolor[HTML]{\ANNColor}4.95e-5 &\cellcolor[HTML]{\ANNColor}&
\cellcolor[HTML]{\ANNColor}0.433 & \cellcolor[HTML]{\ANNColor}2.56e-2 & \cellcolor[HTML]{\ANNColor}1.46e-4 &\cellcolor[HTML]{\ANNColor}&
\cellcolor[HTML]{\ANNColor}0.528 & \cellcolor[HTML]{\ANNColor}6.40e-2 & \cellcolor[HTML]{\ANNColor}4.70e-4 \\ \hline

$S_1A_4$& 
0.294 & 6.25e-3 & 2.42e-5 &  &
0.433 & 1.25e-2 & 7.14e-5 &  &
0.528 & 3.13e-2 & 2.29e-2 \\ \hline

$S_2A_3$& 
0.294 & 3.05e-3 & 1.18e-5&  &
0.433 & 6.10e-3 & 3.48e-5 &  &
0.528 & 1.53e-2 & 1.12e-4 \\ \hline

$S_3A_2$& 
0.294 & 1.45e-3 & 5.61e-6 &  &
0.433 & 2.90e-3 & 1.66e-5 &  &
0.528 & 7.25e-3  & 5.32e-5 \\ \hline

$S_4A_1$& 
0.294 & 6.5e-4 & 2.52e-6 &  &
0.433 & 1.30e-3 & 7.42e-6 &  &
0.528 & 3.25e-3  & 2.38e-5\\ \hline

$S_5A_0$& 
0.294 & 2.5e-4 & 9.67e-7 &  &
0.433 & 5.0e-4 & 2.86e-6 &  &
0.528 & 1.25e-3 & 9.17e-6 \\ \hline

\end{tabular}%
\vspace{3mm}
\end{table*}

\subsection{Power Dissipation}
Figure \ref{fig:dynamic-power} shows a comparison of power consumption among the SNN, ANN, and various hybrid SNN-ANN architectures. The graph illustrates that the ANN baseline, run on CPU and GPU, exhibits significantly higher dynamic power consumption compared to the SNN baseline deployed on Loihi. Adding SpkConv layers reduces power dissipation across the board, although even a very small ANN still consumes vastly more power than a large SNN. A substantial increase in power consumption is observed in configuration $S_5A_0$ across all intervals, but this is an outlier caused due to the model's increased parameter count as shown in Table \ref{tab:hybrid-archs}. This occurred because the accumulator expands the channel dimension of the data, which is normally reduced back down by consecutive Conv layers. However, the accumulator in configuration $S_5A_0$ is followed by a fully connected layer which does not reduce the overall dimensionality. Overall, our results exhibit that replacing non-spiking layers with spiking ones in a hybrid SNN-ANN architecture can generally lead to decreased power dissipation. Similar to latency, the accumulate operation does not consume a substantial amount of power, being less than 1 mW in all cases, which is further discussed in Section \ref{sec:accumulator}.

\subsection{Energy Consumption}
Figure \ref{fig:energy} provides a comparison between the ANN and SNN baselines and the hybrid SNN-ANN models. The results demonstrate that SNNs exhibit significantly lower energy consumption for computational tasks compared to conventional ANNs, highlighting the potential benefits of incorporating spiking layers to reduce energy consumption. It shows that the incremental addition of SpkConv layers significantly decreases energy consumption compared to ANN models. However, there is a similar increase in energy consumption for model $S_5A_0$ due to the model's aforementioned parameter increase. In our experiments, we found that larger accumulate intervals lead to decreased energy consumption across the board. As shown in Table \ref{tab:accumulator-results}, the accumulator only consumes a marginal amount of energy compared to the entire system.

\subsection{Accumulator Overheads}
\label{sec:accumulator}
To accurately estimate the overheads of the accumulator when used in a heterogeneous SoC comprising neuromorphic and ANN accelerator cores, we implemented the accumulator in the Verilog Hardware Description Language (HDL). 

The accumulator circuit includes $k$-bit counters with three input ports corresponding to the input clock, reset, and spike signals. The clock signal is dependent on the hardware system's global clock and the reset signal is driven by the accumulator module. The \textit{spike} signal corresponds to whether a spike was emitted by the SNN during the current clock. Thus, if the last layer of the SNN, connected to the accumulator, spikes, the spikes are transmitted to the accumulator's counters which then increments the $k$-bit counter. 

The high-level \textit{accumulator module} combines multiple copies of the $k$-bit counters into a single hardware accumulator block. The accumulator also has three inputs including the clock signal along with a \textit{sync} signal for synchronizing all of the counters and $N$ input wires to propagate the spike signals to their respective counters. The accumulator also contains an internal $k$-bit register for controlling when the $N$ $k$-bit counters should be reset after a predefined number of clock cycles have passed. The output of the accumulator module consists of $N \times k$ bits which can then be connected to an ANN accelerator core for the ANN inference phase of the hybrid model.

Here, we fix the number of inputs into the accumulator to $N=128$ bits which are then fed into the 128 $k$-bit counters inside of the accumulator. The number of neurons connected to the accumulator from the SNN part of the model often exceeds the limit of the $N=128$. To accommodate these larger layer sizes, we divide the total number of output neurons from the last layer into partitions with 128 neurons and send each partition to the accumulator per timestep. Depending on the value of $k$, the accumulator's output bus would then be $128 \times k$ bits wide. The value of $k$ depends on the accumulate interval (I) value. For example, within the $I=5$ interval, there will be a maximum of 5 spikes. Thus, a 3-bit counter would adequately store the accumulator's output. Similarly, 4-bit and 5-bit counters can support accumulate intervals of $I=10$ and $I=25$, respectively.

% The number of neurons connected to the accumulator often exceeds the limit of the $N=128$ inputs the hardware accumulator accepts. To accommodate these larger layer sizes, we partitioned the computation using the following equation:

% \begin{equation}
%     Partitions = \ceil[\bigg]{\frac{ C \times W \times H }{N}
% \end{equation}

% Here, we divide the total number of output parameters from the last layer per timestep by the number of inputs $N$ into the accumulator. In this case, $N$ is statically set to 128 and the resulting number of partitions is shown in Table \ref{tab:hardware-acc}. Each partition is then fed to the accumulator separately where the spikes are accumulated for the 

Using the Synopsys Design Compiler, we assessed the performance of our accumulator operation on hardware. The findings, detailed in Table \ref{tab:accumulator-results}, indicate that the latency, power, and energy overheads attributed to the accumulator circuit are significantly lower, by several orders of magnitude, compared to those of the ANN and SNN models running on the ANN accelerator and neuromorphic cores. This implies that the accumulator overheads are negligible.

\section{Conclusion}
\label{sec:conclusion}

In this work, we presented a methodology for the efficient deployment of hybrid spiking models on a distributed system of neuromorphic hardware and edge AI accelerators. Our experiments involved testing numerous hybrid models to explore the effect that introducing spiking layers would have on performance. We trained our models on the event-based DvsGesture dataset to perform Gesture Recognition. The models were then profiled on separate devices to record power, latency, and energy. Our findings show that hybrid networks reduce energy consumption compared to homogeneous networks with minimal accuracy loss. We also show that hybrid networks work best with only a few spiking layers serving as feature extractors for following convolution blocks. However, due to the lack of physical hardware, we could not record the cost of communication between devices. This cost would affect latency and energy consumption, potentially offsetting the benefit of hybrid networks. Despite this, the hybrid networks show promising results that can be explored in further research.

%The inclusion of only a few spiking layers caused the hybrid networks to outperform the ANN baseline in every metric. Overall, these hybrid networks show great potential in their utility and viability, combining the significant energy savings of SNNs with the high precision and accuracy of ANNs to perform event-based edge AI applications. %such as sign language recognition and object tracking.

\section*{Acknowledgment}
\small This work is supported by the National Science Foundation (NSF) under grant number 2340249. Special thanks to Intel Labs for providing access to the Loihi chips for the experiments performed in this paper. 

\balance
\bibliographystyle{ieeetr}
\bibliography{refs}

\end{document}